\documentclass{article}

\PassOptionsToPackage{numbers,compress}{natbib}

\usepackage[preprint]{neurips_2026}

\usepackage[utf8]{inputenc}
\usepackage[T1]{fontenc}
\usepackage{hyperref}
\usepackage{url}
\usepackage{booktabs}
\usepackage{amsfonts}
\usepackage{nicefrac}
\usepackage{microtype}
\usepackage{xcolor}
\usepackage{multirow}
\usepackage{amsmath}
\usepackage{amssymb}
\usepackage{graphicx}
\usepackage{colortbl}
\usepackage{wrapfig}
\usepackage{algorithm}
\usepackage{algpseudocodex}
\usepackage{enumitem}
\usepackage{longtable}
\usepackage{placeins}

\definecolor{myframepink}{RGB}{72,138,176}
\hypersetup{colorlinks=true,allcolors=myframepink}

\title{Knowledge Distillation Must Account for What It Loses}

\author{
Wenshuo Wang \\
School of Future Technology, South China University of Technology, China \\
\texttt{202364870251@mail.scut.edu.cn}
}

\begin{document}

\maketitle

\begin{abstract}
This position paper argues that knowledge distillation must account for what it loses: student models should be judged not only by retained task scores, but by whether they preserve the teacher capabilities that make those scores reliable. This matters because distillation is increasingly used to turn large teacher models into deployable students, yet headline metrics can obscure losses in the capabilities that make teacher behavior reliable. Conceptually, we show that current evaluation often assumes retained task scores imply retained teacher capabilities. Reframing distillation as a \textit{lossy projection} exposes this flaw: students may match selected teacher observables without preserving the capabilities that make them reliable. We then synthesize existing evidence into a taxonomy of off-metric distillation losses, showing that such losses are concrete, recurring, and measurable, yet often unaccounted for when studies report what students retain rather than what they lose. To make the position actionable, we propose scenario-specific preservation targets and a Distillation Loss Statement that reports what was preserved, what was lost, and why the remaining losses are acceptable. The goal is not lossless distillation, but accountable distillation.
\end{abstract}

\section{Introduction}\label{sec:intro}

Knowledge distillation transfers behavior from a larger or more capable teacher model to a smaller, cheaper, specialized, or more deployable student model \citep{hinton2015distilling,gou2021survey}. It is now used far beyond classical model compression: modern distillation produces compact language models, reasoning students, code and tool-using agents, retrieval-augmented components, safety models, domain assistants, and synthetic-data pipelines \citep{sanh2019distilbert,xu2024surveyllmkd,deepseek2025r1,kang2025agentdistillation}. These systems are usually evaluated by what the student retains on a primary metric, such as accuracy, pass@k, win rate, benchmark score, task success, or downstream utility. Yet most distillation research does not explicitly state what the student loses beyond the capability represented by that metric. A student can preserve the teacher's headline score while losing other capabilities that make the teacher reliable.

This omission matters because distilled models are increasingly deployed as substitutes for larger systems, not merely studied as compressed approximations in closed benchmarks. In deployment, reliability often depends on capabilities that primary metrics only weakly measure: uncertainty, boundary behavior, process reliability, on-policy stability, grounding, privacy, safety, and diversity. \textbf{This position paper argues that knowledge distillation must account for what it loses.} We do not claim that distillation should be lossless, or that every student must preserve every teacher capability. We argue that distillation papers should identify which teacher capabilities matter for the intended use, measure whether the student preserves them when possible, and justify the losses they accept.

This argument begins by making explicit the \emph{retention assumption} behind current distillation evaluation: that if a student matches the teacher on the primary metric, then it has preserved the relevant teacher capability. As developed in Section~\ref{sec:retention}, this assumption is convenient, but it is not warranted. Matching the teacher under one observable does not imply matching the teacher under others. We therefore reframe distillation as a \emph{lossy projection} of teacher behavior rather than a faithful copy of teacher capability. This framing shifts the central question from how much score the student retained to which teacher capabilities were projected away.

The case for loss accounting is not merely conceptual. As Section~\ref{sec:evidence} shows, prior work has already identified teacher--student divergences in predictive distributions, internal representations, robustness, calibration, subgroup behavior, privacy and memorization, reasoning faithfulness, on-policy stability, refusal boundaries, grounding, and synthetic-data diversity \citep{stanton2021doeskd,ojha2023whatknowledge,mohanty2023whatislost,hebbalaguppe2024calibrationtransfer,stacey2024distillingrobustness,jagielski2023students,zhang2025membershipmemorization,chen2025reasoningdontalways,song2026onpolicy,shumailov2024collapse}. We synthesize these findings into a taxonomy of \emph{off-metric distillation losses}, showing that the problem is not a lack of evidence about teacher--student divergence, but the absence of a reporting norm that accounts for such divergence as loss. Building on that taxonomy, Section~\ref{sec:scenario-preservation} turns this position into a recommendation: distillation should be evaluated through scenario-specific preservation targets, not only by score retention. To make this operational, we propose a \textbf{Distillation Loss Statement} that reports the deployment context, primary metric, critical off-metric capabilities, preservation targets, stress distributions, observed losses, accepted losses, and deployment implications.

Finally, we situate this proposal in Section~\ref{sec:related} relative to adjacent work and objections. Existing research studies particular losses, particular preservation methods, and broader documentation norms, but it does not yet provide a general norm for reporting teacher-to-student capability change \citep{mitchell2019modelcards,gebru2021datasheets,zhao2024measurediversity,tramer2024dp}. The main limitation, revisited in Section~\ref{sec:conclusion}, is that many off-metric capabilities are difficult to define and measure, especially for black-box teachers, and some teacher behaviors should not be preserved. Even so, explicit loss accounting could change how distillation results are interpreted: it would help researchers, reviewers, and deployers distinguish useful compression from unexamined capability erosion, and move distillation from a score-retention exercise toward an accountable transformation.

\section{The Retention Assumption}\label{sec:retention}

This section isolates the assumption that makes score-centered distillation evaluation look sufficient. The issue is not that primary metrics are useless. They are often necessary. The issue is that they are treated as evidence for a broader claim than they can support: that the student has preserved the teacher capability relevant to the intended use.

\subsection{The Hidden Assumption in Current Distillation Evaluation}\label{sec:hidden-assumption}

Let \(T\) denote a teacher model and \(S\) denote a distilled student model. Let \(m(\cdot)\) be the primary evaluation metric used in a distillation paper, such as accuracy, pass rate, win rate, task success, or downstream utility. A common success claim can be abstracted as:
\begin{equation}
    m(S) \approx m(T).
    \label{eq:score-retention}
\end{equation}
Equation~\ref{eq:score-retention} is a claim about retained performance under one metric. It is not, by itself, a claim about retained capability. The hidden assumption appears when this metric-level statement is treated as if it implied a broader teacher--student equivalence.

To make the assumption explicit, let \(c_i(\cdot)\) denote a teacher capability that may matter for deployment but is not directly measured by \(m\). Examples include calibration, refusal boundary, retrieval behavior, self-verification, or long-tail coverage. Let \(\mathcal{C}=\{c_1,\ldots,c_k\}\) be the set of such scenario-critical capabilities. The retention assumption is:
\begin{equation}
    m(S) \approx m(T)
    \quad \Longrightarrow \quad
    c_i(S) \approx c_i(T)
    \quad \text{for relevant } c_i \in \mathcal{C}.
    \label{eq:retention-assumption}
\end{equation}
This implication is not logically warranted. A scalar benchmark can show that two models reach similar outcomes on a test distribution, while leaving open whether they assign similar probabilities, fail on similar inputs, abstain under similar uncertainty, use tools in similar ways, cite similar evidence, or preserve similar minority and tail behavior.

The same problem applies even when the training objective is richer than the final metric. Let \(x\) be an input sampled from a distillation distribution \(\mathcal{D}_{\mathrm{distill}}\). Let \(\phi_a(M,x)\) denote an observable teacher signal of model \(M\) on input \(x\), such as an answer, logit vector, rationale, or trajectory. A distillation objective can be written abstractly as:
\begin{equation}
    \min_{S} \; \mathbb{E}_{x \sim \mathcal{D}_{\mathrm{distill}}}
    \left[d_a\big(\phi_a(T,x),\phi_a(S,x)\big)\right],
    \label{eq:distillation-objective}
\end{equation}
where \(d_a\) is a discrepancy measure for the observed signal. Equation~\ref{eq:distillation-objective} only constrains what is observed, optimized, and evaluated. If the relevant off-metric capability is not represented in \(\phi_a\), or not tested on the distribution where it matters, then retention of the primary metric cannot establish retention of that capability.

\subsection{Distillation as Lossy Projection}\label{sec:lossy-projection}

The retention assumption becomes easier to see if distillation is viewed as a projection. For an input \(x\), let \(\Phi(M,x)\) denote a vector of teacher-relevant observables of model \(M\):
\begin{equation}
    \Phi(M,x)
    =
    \big(\phi_m(M,x),\phi_{c_1}(M,x),\ldots,\phi_{c_k}(M,x)\big),
    \label{eq:capability-vector}
\end{equation}
where \(\phi_m\) is the observable associated with the primary metric and \(\phi_{c_i}\) is an observable or proxy associated with capability \(c_i\). In practice, distillation observes a subset of this vector. Let \(P_A\) be the projection onto the subset of observables \(A\) used for training or evaluation. Distillation tries to make:
\begin{equation}
    P_A\Phi(S,x) \approx P_A\Phi(T,x)
    \quad \text{for } x \sim \mathcal{D}_{\mathrm{distill}}.
    \label{eq:projected-match}
\end{equation}
But matching a projection is not the same as matching the full capability vector:
\begin{equation}
    P_A\Phi(S,x) \approx P_A\Phi(T,x)
    \quad \not\Rightarrow \quad
    \Phi(S,x) \approx \Phi(T,x).
    \label{eq:projection-not-copy}
\end{equation}
This is the sense in which distillation is a \textit{lossy projection}. The projected dimensions may be faithfully transferred, while unprojected dimensions are unconstrained, weakly constrained, or evaluated only indirectly.

We call the resulting discrepancy an \textit{off-metric distillation loss}. For a capability \(c\), let \(\mathcal{D}_c\) be the stress distribution on which that capability matters, let \(\phi_c\) be an observable or proxy for the capability, and let \(d_c\) be a capability-specific discrepancy measure. Then:
\begin{equation}
    \Delta_c(T,S;\mathcal{D}_c)
    =
    \mathbb{E}_{x \sim \mathcal{D}_c}
    \left[d_c\big(\phi_c(T,x),\phi_c(S,x)\big)\right].
    \label{eq:offmetric-loss}
\end{equation}
A student can therefore satisfy the usual score-retention criterion while still incurring a large off-metric loss:
\begin{equation}
    m(S) \approx m(T)
    \quad \text{while} \quad
    \Delta_c(T,S;\mathcal{D}_c) \gg 0.
    \label{eq:score-loss-gap}
\end{equation}
Equation~\ref{eq:score-loss-gap} is the central problem this paper addresses. It does not imply that all off-metric losses are unacceptable, or that every teacher behavior should be preserved. It implies that losses relevant to the intended use should be named, measured when possible, and justified rather than left invisible behind retained benchmark scores.

\section{Evidence for Off-Metric Distillation Loss}\label{sec:evidence}

Section~\ref{sec:retention} showed why primary-metric retention is insufficient as a matter of logic. This section shows why the gap is practically important: we first review what representative distillation work treats as evidence of success in Section~\ref{sec:audit}, then synthesize evidence that teacher--student divergence outside the main metric is not hypothetical in Section~\ref{sec:existing-evidence}.

\subsection{Reporting Patterns in Recent Distillation Work}\label{sec:audit}

A loss-accounting view begins with a simple reporting question: when a paper claims that a student model has successfully inherited a teacher's capability, what evidence is actually reported? In representative distillation work, the default evidence is retention of a primary score, often accompanied by parameter count, latency, or training cost. This is natural in classical compression and remains natural in modern LLM distillation, reasoning distillation, and agent distillation \citep{hinton2015distilling,sanh2019distilbert,xu2024surveyllmkd,deepseek2025r1,kang2025agentdistillation}. The problem is not that such metrics are irrelevant, nor that prior papers report false results. The problem is narrower: primary metrics make some losses visible while leaving other losses outside the reported scope.

We therefore distinguish between \emph{retention evidence} and \emph{loss evidence}. Retention evidence asks whether the student remains competitive on the main task. Loss evidence asks whether the student still matches the teacher on capabilities that the main task does not measure: predictive-distribution fidelity, calibration, robustness, subgroup behavior, process reliability, on-policy stability, grounding, safety boundaries, privacy, and tail diversity. The literature treats many of these capabilities as specialized topics rather than default reporting items \citep{shao2022adversarial,hebbalaguppe2024calibrationtransfer,mohammadshahi2025leftafter,jagielski2023students,zhang2026responsebasedkd}. This pattern is itself informative. Off-metric capabilities are usually treated as specialized research topics, not as default reporting items in ordinary distillation papers. Appendix~\ref{app:reporting-checklist} provides a representative 50-paper reporting-pattern checklist that makes this asymmetry explicit without treating it as a systematic meta-analysis. Our position is that this separation should narrow: papers need not report every possible loss, but they should report the losses that matter for the student's intended use.

\subsection{Existing Evidence That Primary-Metric Retention Is Insufficient}\label{sec:existing-evidence}

The evidence is best read as a set of parallel examples of the same pattern: a student can retain the primary score while diverging from the teacher along dimensions that the primary metric does not measure. We organize these examples by the kind of off-metric loss they make visible.

\paragraph{Predictive-distribution loss.}
The original distillation formulation already treats soft targets as carrying information beyond hard labels \citep{hinton2015distilling}. Later analyses make the point sharper: teacher probability estimates can matter even when teacher accuracy is not the only object of interest, and students can improve under distillation while still failing to match teacher predictive distributions \citep{menon2021statistical,stanton2021doeskd}. Recent LLM distillation methods likewise show that the direction and form of distribution matching matter for generation quality \citep{gu2024minillm}. These results support a narrow but important conclusion: answer-level or score-level similarity is weaker than distribution-level fidelity.

\paragraph{Representation and relation loss.}
Representation- and relation-based distillation methods exist precisely because matching final responses may not preserve intermediate structure. FitNets use teacher hints from intermediate layers, while MiniLM transfers self-attention distributions and value relations; broader KD surveys distinguish response-based, feature-based, and relation-based knowledge for the same reason \citep{romero2015fitnets,wang2020minilm,gou2021survey}. Empirical studies of what knowledge gets distilled also show that properties such as invariances, localization behavior, and adversarial susceptibility can be inherited unevenly \citep{ojha2023whatknowledge,mohanty2023whatislost}. The lesson is not that every representation must be copied. It is that output retention alone cannot tell us which internal or relational properties were preserved.

\paragraph{Capability-boundary loss.}
Average-case accuracy can remain high while robustness, group behavior, or tail behavior changes. Work on adversarial robustness transfer argues that standard KD is usually evaluated by accuracy while robustness may fail to transfer \citep{shao2022adversarial}. In natural language inference, in-distribution gains do not automatically imply robustness on target or out-of-distribution examples \citep{stacey2024distillingrobustness}. Work on bias and fairness shows that the effects of KD can be uneven across subgroups or classes \citep{lukasik2021teacherspet,mohammadshahi2025leftafter}. These results motivate the notion of a capability boundary: the student may preserve the center of the teacher's behavior while distorting the regions where reliability is most contested.

\paragraph{Privacy and memorization loss.}
Distillation is sometimes treated as privacy-improving because the student interacts with private training data only indirectly, yet membership-inference work shows that distillation alone can provide limited privacy protection \citep{jagielski2023students}. Recent LLM studies further show that students can inherit membership and memorization risks from teachers, while different KD objectives can induce different leakage profiles \citep{zhang2025membershipmemorization,borkar2026memorization}. The relevant point is not that distillation always worsens privacy. It is that distillation changes the privacy and memorization profile of the student, and this change is invisible under ordinary task scores.

\paragraph{Calibration and abstention loss.}
Calibration research shows that confidence reliability is distinct from accuracy, and calibration transfer studies show that this property is not automatically inherited by a student \citep{guo2017calibration,hebbalaguppe2024calibrationtransfer,fan2024revisitcalibration}. In LLM settings, uncertainty estimation and abstention are increasingly treated as independent capabilities: a model must know when it does not know, and when it should decline or defer rather than answer \citep{geng2024surveyconfidence,kapoor2024knowwhat,wen2024abstention}. If a student retains a teacher's answer style but not its uncertainty behavior, then the student may become more deployable while becoming less trustworthy.

\begin{table}[t]
\centering
\caption{A taxonomy of off-metric distillation losses. Each row identifies a teacher capability that may be weakly measured or unmeasured by the primary score. The table should be read as a loss-accounting checklist, not as a claim that all losses occur in every distillation setting.}
\label{tab:offmetric-loss-taxonomy}
\resizebox{\linewidth}{!}{%
\begin{tabular}{p{0.19\linewidth}p{0.22\linewidth}p{0.34\linewidth}p{0.25\linewidth}}
\toprule
\textbf{Off-metric loss} & \textbf{What the primary metric may retain} & \textbf{What the student may lose} & \textbf{Representative evidence} \\
\midrule
Predictive-distribution loss & Top answer or task score & Soft probabilities, alternative answers, entropy, teacher uncertainty shape & Soft targets and distribution fidelity \citep{hinton2015distilling,menon2021statistical,stanton2021doeskd,gu2024minillm} \\
\addlinespace
Representation / relation loss & Similar output behavior & Intermediate representations, attention relations, invariances, localization behavior & Feature-, relation-, and property-transfer studies \citep{romero2015fitnets,wang2020minilm,ojha2023whatknowledge,mohanty2023whatislost} \\
\addlinespace
Capability-boundary loss & Average-case accuracy or win rate & OOD robustness, adversarial robustness, subgroup behavior, class-wise reliability & Robustness and fairness studies \citep{shao2022adversarial,stacey2024distillingrobustness,lukasik2021teacherspet,mohammadshahi2025leftafter} \\
\addlinespace
Privacy / memorization loss & Task utility or reduced average memorization & Membership leakage, teacher-specific memorized examples, different leakage profiles across KD objectives & Privacy and memorization studies \citep{jagielski2023students,zhang2025membershipmemorization,borkar2026memorization} \\
\addlinespace
Calibration / abstention loss & Correct answers on answer-forcing benchmarks & Confidence reliability, knowing when not to answer, deferral or abstention behavior & Calibration and abstention work \citep{guo2017calibration,hebbalaguppe2024calibrationtransfer,fan2024revisitcalibration,kapoor2024knowwhat,wen2024abstention} \\
\addlinespace
Process / on-policy stability loss & Final answer, plausible rationale, or teacher-forced trace & Search behavior, self-verification, recovery from self-generated errors, rollout stability, faithful reasoning traces & CoT, process-faithfulness, and on-policy studies \citep{hsieh2023dss,yu2024distills2tos1,turpin2023dontalways,chen2025reasoningdontalways,song2026onpolicy} \\
\addlinespace
Grounding / provenance loss & Factual-looking answer text & Retrieval policy, evidence selection, citation fidelity, no-answer behavior under missing evidence & RAG and attribution work \citep{lewis2020rag,jia2025rationaledistrag,huang2024groundedcitations} \\
\addlinespace
Safety-boundary loss & Refusal style or aggregate safety score & Jailbreak robustness, over-refusal / under-refusal boundary, grounded selective refusal & Refusal and jailbreak studies \citep{cui2024orbench,muhamed2026refusalbench,zhang2026responsebasedkd} \\
\addlinespace
Long-tail / diversity loss & High-frequency patterns and downstream average score & Rare modes, minority classes, human heterogeneity, tail coverage in generated data & Model-collapse and diversity work \citep{shumailov2024collapse,gerstgrasser2024inevitable,zhao2024measurediversity} \\
\bottomrule
\end{tabular}%
}
\end{table}

\paragraph{Process, grounding, safety, and diversity loss.}
Modern LLM applications expose additional losses that are weakly captured by headline metrics. Step-by-step and System-2-to-System-1 distillation demonstrate that reasoning traces and deliberative procedures can be compressed, but work on chain-of-thought faithfulness shows that explanations need not faithfully reveal the model's actual decision process \citep{hsieh2023dss,yu2024distills2tos1,turpin2023dontalways,chen2025reasoningdontalways}. On-policy distillation work makes a related point for generative students: training on static teacher-generated data can leave students untested on their own errors, creating exposure bias and compounding failures under student-generated rollouts \citep{song2026onpolicy}. Retrieval and citation work shows that answer quality can diverge from evidence selection and attribution quality \citep{lewis2020rag,jia2025rationaledistrag,huang2024groundedcitations}. Safety work shows that refusal behavior must be evaluated at the boundary between over-refusal and under-refusal, not merely by the presence of refusal text \citep{cui2024orbench,muhamed2026refusalbench,zhang2026responsebasedkd}. Recursive data-generation work shows that generated-data pipelines can lose tails of the original distribution, even when average performance appears acceptable \citep{shumailov2024collapse,gerstgrasser2024inevitable}. Together, these lines of work support the same conclusion: the main metric does not exhaust what the teacher knows how to do.

\paragraph{Single-scenario loss accounting.}
Existing single-scenario studies instantiate the kind of loss accounting we call for. For example, metamorphic testing for distilled code models shows that students can preserve conventional accuracy while failing to deeply mimic teacher behavior under behavior-preserving transformations; MetaCompress reports up to a 285\% greater performance drop under adversarial attacks and identifies up to 62\% behavioral discrepancies that accuracy-based evaluation misses \citep{awal2025metacompress}. This is not a general solution, because it is tailored to code models and metamorphic relations. It is instead an existence proof for our central claim: measuring what the student retains is insufficient unless we also ask what the student loses.

\subsection{A Taxonomy of Off-Metric Distillation Losses}\label{sec:loss-taxonomy}

Table~\ref{tab:offmetric-loss-taxonomy} organizes the evidence above into a loss-oriented taxonomy. The taxonomy is not intended to be exhaustive, and it does not imply that every distillation paper must measure every row. Instead, each row names a capability dimension that can be invisible under the primary metric and suggests the kind of evidence that would make the loss visible.

The table also clarifies how the formal loss in Equation~\ref{eq:offmetric-loss} should be instantiated. For each row, a paper must choose a capability \(c\), an observable or proxy \(\phi_c\), a stress distribution \(\mathcal{D}_c\), and a discrepancy measure \(d_c\). For calibration, \(\phi_c\) might be a confidence estimate and \(\mathcal{D}_c\) might contain ambiguous or shifted inputs. For grounding, \(\phi_c\) might be retrieved evidence and citation alignment. For safety, \(\mathcal{D}_c\) might be a set of boundary prompts rather than ordinary helpfulness prompts. For privacy, \(\phi_c\) might be a membership-inference or memorization probe. This is why loss accounting must be scenario-specific. The relevant losses are not determined by distillation in the abstract; they are determined by what the student is expected to do.

\section{From Loss Taxonomy to Scenario-Specific Preservation}\label{sec:scenario-preservation}

\begin{table}[t]
\centering
\caption{Scenario-specific preservation targets for distillation. The primary metric remains useful, but it should be paired with preservation targets for the off-metric capabilities that determine whether the distilled student is reliable in the intended setting.}
\label{tab:scenario-preservation}
\resizebox{\linewidth}{!}{%
\begin{tabular}{p{0.20\linewidth}p{0.23\linewidth}p{0.29\linewidth}p{0.34\linewidth}}
\toprule
\textbf{Distillation scenario} & \textbf{Typical primary metric} & \textbf{Capabilities to account for} & \textbf{Preservation targets and stress tests} \\
\midrule
Reasoning-model students & Math, science, coding, or reasoning benchmarks; pass rate; self-consistency score & Capability boundary, self-verification, process reliability, when to spend more inference compute & Multi-sample teacher behavior, verifier scores, failed traces, hard and near-boundary problems, tests of reasoning faithfulness \\
\addlinespace
Code and tool-using agents & Task success, resolved issue rate, pass@k, patch correctness & Debugging trajectory, test generation, tool-use policy, recovery from failed attempts, on-policy stability & Tool traces, execution logs, failed patches, student-generated rollouts, ambiguous specifications, hidden-test stress cases \\
\addlinespace
RAG and domain QA systems & Exact match, F1, factuality, answer preference & Retrieval policy, evidence selection, citation fidelity, no-answer behavior under missing evidence & Retrieved documents, selected evidence, attribution alignment, insufficient-evidence cases, citation-support checks \\
\addlinespace
High-risk domain assistants & Expert QA accuracy, preference, case-level correctness & Calibration, abstention, scope boundary, escalation or referral behavior & Confidence and uncertainty estimates, abstention labels, risk categories, incomplete-context cases, shifted or ambiguous cases \\
\addlinespace
Privacy-sensitive students & Task utility, compression, or retained benchmark score & Membership leakage, memorization, teacher-specific example inheritance, compliance-relevant data exposure & Membership-inference probes, exposure or canary tests, memorization probes, comparisons across soft and hard KD objectives \\
\addlinespace
Safety and refusal students & Aggregate safety score, jailbreak success rate, helpfulness & Refusal boundary, over-refusal / under-refusal tradeoff, robustness to boundary prompts & Near-boundary unsafe requests, benign hard requests, multilingual jailbreaks, selective-refusal evaluations \\
\addlinespace
Synthetic-data pipelines & Downstream accuracy, label agreement, generation cost & Tail coverage, diversity, minority modes, human heterogeneity & Rare-class coverage, diversity measures, human-anchor ratios, comparisons to real-data tails \\
\addlinespace
On-device or local students & Retained benchmark score, latency, memory, throughput & Safety margin, calibration, multi-turn consistency, graceful degradation under limited compute & Resource-constrained stress tests, calibration probes, safety boundary tests, long-context or multi-turn probes \\
\bottomrule
\end{tabular}%
}
\end{table}

The taxonomy in Section~\ref{sec:loss-taxonomy} is useful only if it changes what distillation papers choose to preserve and report. It should not be read as a universal checklist. A reasoning student, a RAG component, a privacy-sensitive assistant, and an on-device classifier need not preserve the same teacher capabilities. The central requirement is narrower and more practical: for each intended use, authors should identify which off-metric capabilities are consequential, choose preservation targets for them, and evaluate the student on stress distributions where those capabilities matter.

This shifts distillation from a generic score-retention problem to a scenario-specific preservation problem. In the notation of Section~\ref{sec:retention}, a paper should not only report whether \(m(S) \approx m(T)\), but also choose a small set of relevant capabilities \(c \in \mathcal{C}\), observable proxies \(\phi_c\), stress distributions \(\mathcal{D}_c\), and discrepancy measures \(d_c\). The goal is not to make every student imitate every teacher property. It is to make explicit which properties are being preserved, which are being ignored, and why that choice is appropriate for the deployment context.

\subsection{What Common Distillation Scenarios Should Preserve}\label{sec:scenario-targets}

Table~\ref{tab:scenario-preservation} gives representative preservation targets for common distillation settings. The table is intentionally phrased as guidance rather than a benchmark specification. It connects each scenario to the capability losses most likely to matter beyond the primary metric, and to the kinds of teacher signals or stress tests that would make those losses visible.

Several examples illustrate the logic. Reasoning distillation can transfer rationales or compress deliberative procedures, but this does not by itself establish that the student preserves faithful reasoning, self-verification, or the boundary between problems it can and cannot solve \citep{deepseek2025r1,hsieh2023dss,yu2024distills2tos1,turpin2023dontalways,chen2025reasoningdontalways}. Agent distillation should therefore preserve more than the final answer: tool calls, failed attempts, tests, recovery behavior, and stability under student-generated rollouts can be part of the teacher capability that matters \citep{kang2025agentdistillation,song2026onpolicy}. RAG distillation similarly should not reduce evidence-grounded behavior to answer text; retrieval, attribution, and insufficient-evidence decisions are separate preservation targets \citep{lewis2020rag,jia2025rationaledistrag,huang2024groundedcitations}. Privacy-sensitive distillation should not infer privacy from compression alone, because membership leakage and memorization can change with the KD objective \citep{jagielski2023students,zhang2025membershipmemorization,borkar2026memorization}. In safety, the relevant object is not refusal style but the boundary between appropriate refusal and over-refusal \citep{cui2024orbench,muhamed2026refusalbench,zhang2026responsebasedkd}. In synthetic-data pipelines, the crucial loss may be diversity and tail coverage rather than immediate downstream score \citep{shumailov2024collapse,gerstgrasser2024inevitable,zhao2024measurediversity}. Across these cases, the same principle holds: the preservation target should be chosen from the capability that makes the primary score trustworthy.

\subsection{A Distillation Loss Statement}\label{sec:distillation-loss-statement}

To make scenario-specific preservation actionable, we propose that distillation papers include a \textbf{Distillation Loss Statement}. This statement is a short reporting component, analogous in spirit to documentation practices for models and datasets \citep{mitchell2019modelcards,gebru2021datasheets}, but focused specifically on what changes when a teacher becomes a student. It should be concise enough to be used in ordinary distillation papers, while explicit enough to prevent score retention from being mistaken for capability preservation.

\begin{table}[t]
\centering
\caption{A template for a Distillation Loss Statement. The template does not require every paper to measure every possible loss. It requires authors to state which losses matter for the intended use and how those losses were handled.}
\label{tab:dls-template}
\resizebox{\linewidth}{!}{%
\begin{tabular}{p{0.25\linewidth}p{0.74\linewidth}}
\toprule
\textbf{Item} & \textbf{Question to answer} \\
\midrule
Deployment context & What is the student intended to do, and under what use conditions? \\
Primary metric & Which score is used to claim successful distillation? \\
Critical off-metric capabilities & Which teacher capabilities matter beyond the primary metric in this context? \\
Preservation targets & Which teacher signals or proxies are used to preserve those capabilities? \\
Stress distributions & On which boundary, shifted, ambiguous, adversarial, privacy-sensitive, on-policy, or tail cases is preservation evaluated? \\
Observed or unmeasured losses & Where does the student diverge from the teacher outside the primary metric, and which relevant losses were not measured? \\
Accepted losses or intended divergences & Which losses are considered acceptable, which divergences are corrective rather than harmful, and why? \\
Deployment implication & What should users, reviewers, or deployers infer from the remaining losses? \\
\bottomrule
\end{tabular}%
}
\end{table}

Table~\ref{tab:dls-template} gives the reporting template. A Distillation Loss Statement could be written in a compact form:
\begin{quote}
\emph{This student is intended for [deployment context]. Its primary metric is [metric]. The critical off-metric teacher capabilities are [capabilities]. We attempt to preserve them using [teacher signals or proxies] and evaluate them on [stress distributions]. The student diverges from the teacher in [observed losses]. We consider these losses [acceptable or unacceptable], or these divergences [intended or corrective], because [justification], with the following deployment implications: [implications].}
\end{quote}

We intentionally do not prescribe universal quantitative thresholds for acceptable loss. Risk-management and documentation frameworks emphasize that evaluation and acceptable use should be tied to intended use, limitations, and deployment context rather than treated as context-free properties of a model \citep{mitchell2019modelcards,gebru2021datasheets,tabassi2023airmf}. A calibration gap that is acceptable for a toy classifier may be unacceptable for a medical assistant; a refusal-boundary shift that is acceptable for a creative-writing assistant may be unacceptable for a safety-critical deployment; a memorization profile that is tolerable for a public benchmark model may be unacceptable for a private-data setting. The Distillation Loss Statement therefore standardizes the \textit{question}, not the \textit{answer}. Authors should identify the relevant capability, choose a suitable proxy or metric when one is available, evaluate it on a meaningful stress distribution, and justify the acceptability of the resulting loss in context.

The statement has two purposes. First, it disciplines evaluation: authors must decide whether calibration, grounding, safety boundary, privacy, diversity, process reliability, or some other capability is relevant, rather than implicitly treating the primary score as sufficient. Second, it disciplines interpretation: reviewers and deployers can see whether a distilled model is a faithful substitute, a task-specific approximation, a corrective departure from an imperfect teacher, or a useful but narrower system. This is why the proposal is compatible with successful distillation. A student may be smaller, faster, cheaper, and still worth deploying even if it loses some teacher capabilities. The point is that these losses should be named, measured when possible, and justified rather than hidden behind retained benchmark performance.

\section{Alternative Views and Adjacent Work}\label{sec:related}

\subsection{Alternative Views and Objections}\label{sec:objections}

We consider several viable objections that clarify the scope of accountable distillation: lossy compression, low-risk deployment, measurement burden, limited teacher access, and corrective divergence.

\paragraph{Distillation is already lossy.}
Classical compression and KD were designed to make useful students smaller, cheaper, or faster, not to reproduce every teacher property \citep{bucila2006modelcompression,hinton2015distilling,gou2021survey}. This objection narrows, rather than weakens, our claim. If losses are expected, then consequential losses should be named and justified rather than hidden behind retained benchmark performance.

\paragraph{Primary metrics may suffice in low-risk settings.}
We agree. A narrow task with stable inputs and limited deployment consequences may need only a short loss statement. This is why our proposal is scenario-specific: it follows model, data, and risk-documentation traditions that tie reporting to intended use, limitations, and operating conditions \citep{mitchell2019modelcards,gebru2021datasheets,tabassi2023airmf}.

\paragraph{Loss accounting may be costly or inconsistent.}
The cost concern is real, but universal thresholds would be misleading. Measurement choices must match the construct being measured, and AI risk-management frameworks treat risk tolerance as context-dependent rather than fixed across use cases \citep{zhao2024measurediversity,tabassi2023airmf}. We therefore standardize the reporting question, not the numerical answer: authors should measure consequential losses when suitable proxies exist, disclose when they do not, and justify acceptability in context.

\paragraph{Black-box teachers make perfect accounting impossible.}
In LLM distillation, available teacher signals vary across black-box and white-box settings \citep{xu2024surveyllmkd}. But black-box access still permits behavioral proxies: multi-sample behavior, self-consistency, refusal boundaries, tool traces, retrieval sets, privacy probes, stress distributions, and human or automated evaluations can reveal divergences that final scores miss \citep{wang2023selfconsistency,gu2024minillm,kang2025agentdistillation,jagielski2023students,zhang2025membershipmemorization}. The appropriate demand is explicit measurement of what can be observed and explicit acknowledgment of what cannot.

\paragraph{Not all divergence is loss.}
A student may outperform or intentionally correct its teacher \citep{furlanello2018bornagain}. It may also reduce undesirable teacher behavior, such as over-refusal, unsupported citation, or overconfident prediction \citep{cui2024orbench,huang2024groundedcitations,guo2017calibration}. This is not a problem for loss accounting. Rather, it shows why teacher--student divergence should be classified, not automatically condemned: authors should distinguish harmful loss, acceptable narrowing, unmeasured divergence, and intended correction.

Together, these objections narrow the proposal rather than refute it. We do not argue for exhaustive preservation, metric proliferation for its own sake, or a single benchmark for all distillation settings. We argue for a reporting norm proportional to intended use. The more a student substitutes for a teacher in open, uncertain, privacy-sensitive, or high-stakes settings, the stronger the need to account for capabilities that may have been projected away.

\subsection{Related Work}\label{sec:adjacent-work}

Adjacent work falls into four broad groups. The first studies teacher--student fidelity and asks what distillation actually transfers: predictive distributions, learned invariances, intermediate representations, or other properties beyond the final answer \citep{stanton2021doeskd,ojha2023whatknowledge,mohanty2023whatislost}. The second develops preservation methods or diagnostics for particular capabilities, such as representation, relation, robustness, calibration, fairness, reasoning traces, agent behavior, grounding, refusal behavior, privacy, memorization, on-policy stability, or behavioral fidelity under metamorphic tests \citep{romero2015fitnets,wang2020minilm,shao2022adversarial,hebbalaguppe2024calibrationtransfer,stacey2024distillingrobustness,hsieh2023dss,kang2025agentdistillation,zhang2026responsebasedkd,jagielski2023students,zhang2025membershipmemorization,borkar2026memorization,song2026onpolicy,awal2025metacompress}. The third proposes broader documentation and evaluation norms, such as model cards, datasheets, and position papers calling for more careful measurement of under-specified claims \citep{mitchell2019modelcards,gebru2021datasheets,zhao2024measurediversity,tramer2024dp}. The fourth consists of existing worked instances in which a single domain already measures a loss that ordinary score retention would miss, such as privacy leakage in LLM KD or behavioral discrepancies in distilled code models \citep{zhang2025membershipmemorization,awal2025metacompress}.

These lines of work are complementary to our position, but they do not replace it. Fidelity studies show that students and teachers can diverge, but they do not provide a general reporting norm for intended-use losses. Capability-specific methods show that particular losses can be reduced or diagnosed, but they do not tell authors which losses matter in which deployment scenario. Documentation frameworks show that machine learning artifacts can be reported more responsibly, but they do not focus on the teacher-to-student transformation itself. Existing single-domain loss-accounting studies show that the approach is feasible, but they remain local. Our contribution is to connect these threads into a single claim: distillation should be treated as a lossy transformation whose consequential losses must be accounted for. This claim is not mutually exclusive with prior methods. It explains when those methods are needed, how their results should be interpreted, and why retained score alone should not be the default evidence of successful distillation.

\section{Conclusion}\label{sec:conclusion}

This paper argues that knowledge distillation should be evaluated not only by what students retain on headline metrics, but also by what they lose in teacher capabilities that are critical for deployment. We identified the retention assumption underlying score-centered evaluation, reframed distillation as a \textit{lossy projection}, synthesized evidence of off-metric losses, and proposed scenario-specific preservation targets together with a Distillation Loss Statement. This matters because distillation is increasingly the bridge between frontier-scale models and deployable systems. What is lost at that bridge can determine whether a student is merely smaller or also narrower, less calibrated, less grounded, less privacy-preserving, less stable on its own rollouts, and less safe. If adopted, loss accounting would change how distillation results are interpreted: retained performance would no longer count as sufficient evidence of preserved capability. It would give authors, reviewers, and deployers a vocabulary for distinguishing useful compression from unexamined capability erosion, and could shift distillation research from score retention toward accountable model transformation.

The main limitation of this position is that off-metric capabilities are difficult to define and measure, especially when the teacher is black-box and the relevant capability is only observable through imperfect behavioral proxies. Some teacher behaviors also should not be preserved, because they may encode bias, overconfidence, unsafe refusals, or other undesirable traits. These limitations do not remove the need for loss accounting. They clarify its purpose: distillation papers should not promise lossless transfer, but should make the consequential losses visible, measurable when possible, and justified for the intended use.

\section*{References}
\begingroup
\small
\renewcommand{\section}[2]{}

\endgroup

\appendix
\clearpage

\section{Representative Reporting-Pattern Checklist}\label{app:reporting-checklist}

Table~\ref{tab:reporting-checklist} is a representative 50-paper checklist used to support the reporting-pattern discussion in Section~\ref{sec:audit}. It is not a systematic meta-analysis and does not validate or reject the empirical claims of the cited papers. Instead, it records how the literature is distributed across two kinds of evidence: \emph{retention evidence}, which primarily documents task performance, compression, or deployment efficiency; and \emph{loss evidence}, which makes a teacher--student divergence visible beyond the primary metric. The point is not that every paper should report every row. The point is that the field already has many local tools for measuring particular losses, but lacks a general norm requiring authors to state which losses matter for the intended use.

\small
\begin{longtable}{p{0.05\linewidth}p{0.22\linewidth}p{0.63\linewidth}}
\caption{Representative 50-paper reporting-pattern checklist. The table maps cited work to the kind of evidence it makes salient for this position paper.}\label{tab:reporting-checklist}\\
\toprule
\textbf{Work} & \textbf{Evidence type} & \textbf{Role in our argument} \\
\midrule
\endfirsthead
\toprule
\textbf{Work} & \textbf{Evidence type} & \textbf{Role in our argument} \\
\midrule
\endhead
\bottomrule
\endfoot
\citep{hinton2015distilling} & Retention and distribution & Introduces soft targets as information beyond hard labels. \\
\citep{gou2021survey} & Method taxonomy & Distinguishes response-, feature-, and relation-based KD. \\
\citep{sanh2019distilbert} & Retention evidence & Shows successful compression and retained language-understanding performance. \\
\citep{xu2024surveyllmkd} & LLM KD background & Documents the diversity of modern LLM distillation settings. \\
\citep{deepseek2025r1} & Reasoning distillation & Illustrates the contemporary importance of distilled reasoning students. \\
\citep{kang2025agentdistillation} & Agent/tool distillation & Shows that tool behavior can become a distillation target. \\
\citep{stanton2021doeskd} & Distribution loss & Shows student predictive distributions may diverge from teachers. \\
\citep{menon2021statistical} & Distribution loss & Explains why teacher probability estimates can matter beyond accuracy. \\
\citep{gu2024minillm} & Generative distribution & Studies how distribution-matching choices affect LLM KD. \\
\citep{phuong2019understanding} & Theory & Analyzes why KD can work without reducing success to score retention. \\
\citep{ojha2023whatknowledge} & Property transfer & Studies which off-task properties are inherited by students. \\
\citep{mohanty2023whatislost} & Loss study & Directly studies information loss between teacher and student. \\
\citep{romero2015fitnets} & Representation preservation & Uses intermediate hints, showing outputs alone may be insufficient. \\
\citep{wang2020minilm} & Relation preservation & Transfers attention and value relations, not only final outputs. \\
\citep{furlanello2018bornagain} & Counterpoint & Shows students may outperform teachers on some metrics. \\
\citep{shao2022adversarial} & Robustness loss & Shows adversarial robustness may fail to transfer under KD. \\
\citep{stacey2024distillingrobustness} & OOD loss & Shows in-distribution gains do not guarantee target robustness. \\
\citep{lukasik2021teacherspet} & Subgroup behavior & Studies uneven group-wise effects of distillation. \\
\citep{mohammadshahi2025leftafter} & Fairness loss & Examines fairness and bias after knowledge transfer. \\
\citep{guo2017calibration} & Calibration metric & Establishes confidence calibration as distinct from accuracy. \\
\citep{hebbalaguppe2024calibrationtransfer} & Calibration transfer & Studies whether calibration transfers through KD. \\
\citep{fan2024revisitcalibration} & Calibration as KD & Treats calibration as central to distilling knowledge. \\
\citep{geng2024surveyconfidence} & Uncertainty background & Surveys confidence estimation and calibration in LLMs. \\
\citep{kapoor2024knowwhat} & Uncertainty behavior & Argues models must learn what they do not know. \\
\citep{wen2024abstention} & Abstention & Surveys abstention as a distinct LLM capability. \\
\citep{hsieh2023dss} & Rationale distillation & Shows rationales can improve small-model learning. \\
\citep{wei2022cot} & Reasoning traces & Establishes chain-of-thought as an important reasoning signal. \\
\citep{li2023symboliccot} & CoT distillation & Shows small models can learn step-by-step symbolic rationales. \\
\citep{yu2024distills2tos1} & Process compression & Compresses deliberative procedures into faster students. \\
\citep{lanham2023faithfulness} & Process faithfulness & Measures whether CoT reflects underlying reasoning. \\
\citep{turpin2023dontalways} & Unfaithful rationales & Shows explanations can misrepresent model reasoning. \\
\citep{madsen2024selfexplanations} & Explanation faithfulness & Studies faithfulness of LLM self-explanations. \\
\citep{chen2025reasoningdontalways} & Reasoning faithfulness & Shows reasoning models may not disclose what drives answers. \\
\citep{lewis2020rag} & Grounding architecture & Establishes retrieval-augmented generation as a source-grounded setting. \\
\citep{jia2025rationaledistrag} & RAG distillation & Distills rationales for retrieval-augmented generation. \\
\citep{huang2024groundedcitations} & Citation fidelity & Studies fine-grained grounded citations. \\
\citep{cui2024orbench} & Over-refusal & Evaluates excessive refusal as a distinct failure mode. \\
\citep{muhamed2026refusalbench} & Selective refusal & Evaluates grounded selective refusal. \\
\citep{zhang2026responsebasedkd} & Safety transfer & Shows response-based KD can compromise jailbreak prevention. \\
\citep{cao2023learntorefuse} & Refusal mechanism & Treats refusal as an explicit capability. \\
\citep{shumailov2024collapse} & Tail loss & Shows recursive generated-data training can collapse distribution tails. \\
\citep{gerstgrasser2024inevitable} & Synthetic-data mitigation & Studies when model collapse may be avoided. \\
\citep{zhao2024measurediversity} & Measurement norm & Argues diversity claims require explicit measurement. \\
\citep{mitchell2019modelcards} & Reporting norm & Provides a model documentation precedent. \\
\citep{gebru2021datasheets} & Reporting norm & Provides a dataset documentation precedent. \\
\citep{awal2025metacompress} & Worked loss instance & Uses metamorphic testing to reveal behavioral discrepancies in distilled code models. \\
\citep{jagielski2023students} & Privacy leakage & Shows distillation can provide limited protection against membership inference. \\
\citep{zhang2025membershipmemorization} & LLM privacy loss & Studies membership and memorization risks in LLM KD. \\
\citep{borkar2026memorization} & Memorization dynamics & Shows KD can change memorization and teacher-specific inheritance profiles. \\
\citep{song2026onpolicy} & On-policy stability & Connects static teacher data to exposure bias and student-rollout instability. \\
\end{longtable}
\normalsize

\end{document}